\documentclass[letterpaper]{article}
\usepackage[preprint]{aaai2027}
\usepackage[hyphens]{url}
\usepackage{graphicx}
\usepackage{natbib}
\usepackage{caption}
\usepackage{algorithm}
\usepackage{algorithmic}
\usepackage{amssymb}
\usepackage{amsmath}
\usepackage{comment}
\usepackage{subcaption}
\usepackage{booktabs}
\usepackage{xcolor}
\usepackage{pifont}
\usepackage[export]{adjustbox}

\newcommand*{\inlineimg}[1]{%
  \raisebox{-.22\baselineskip}{%
    \includegraphics[height=1.05\baselineskip,width=1.05\baselineskip,keepaspectratio]{#1}%
  }%
}
\newcommand*{\inlinecard}[1]{%
  \raisebox{-.25ex}{\includegraphics[height=2.1ex,keepaspectratio]{#1}}%
}
\newcommand{\topagent}{\inlineimg{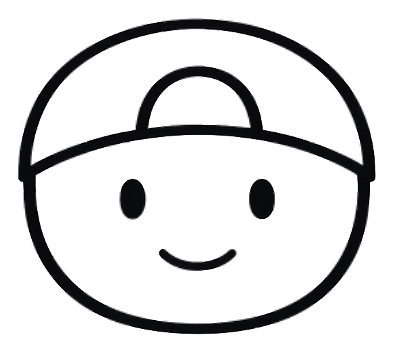}}
\newcommand{\bottomagent}{\inlineimg{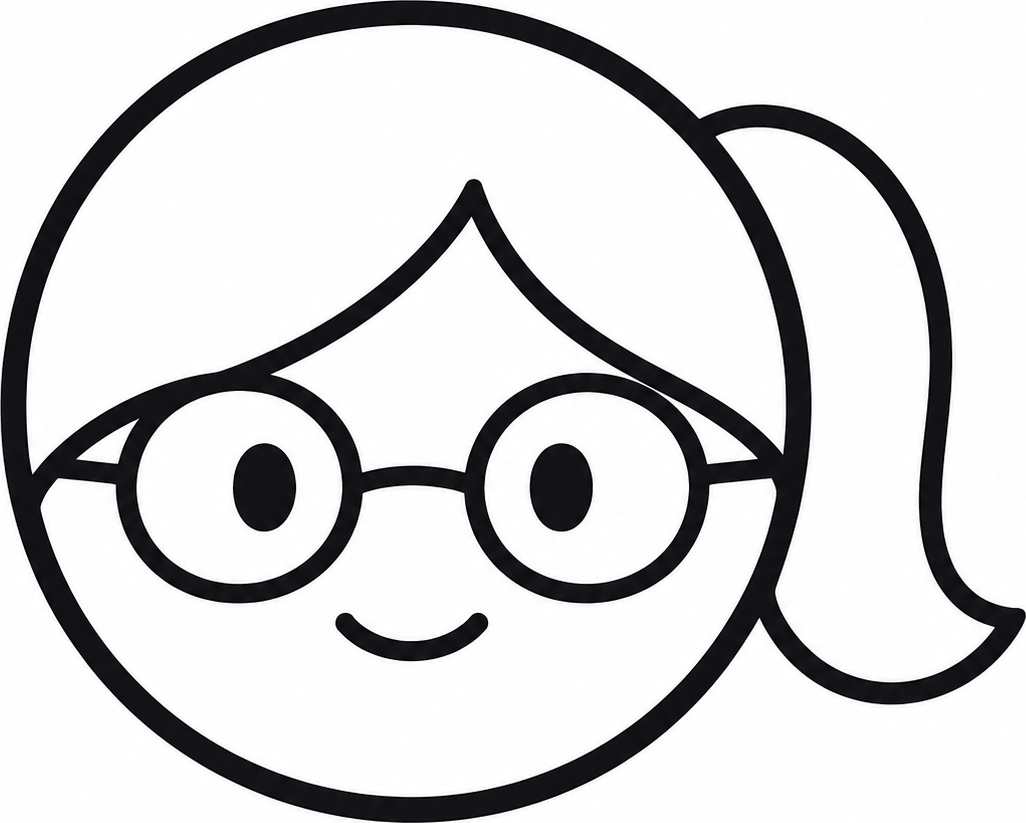}}
\newcommand{\redcard}{\inlinecard{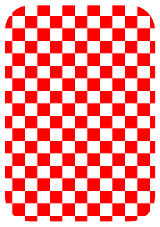}}
\newcommand{\bluecard}{\inlinecard{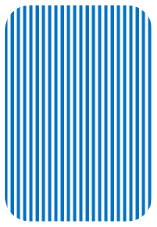}}
\newcommand{\greencard}{\inlinecard{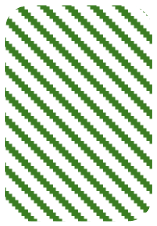}}
\newcommand{\yellowcard}{\inlinecard{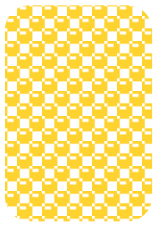}}
\newcommand{\purplecard}{\inlinecard{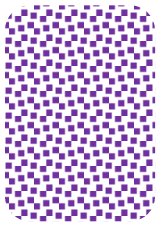}}

\usepackage{etoolbox}
\makeatletter
\patchcmd{\@maketitle}{These authors contributed equally.}{Equal contribution.}{}{}
\patchcmd{\@maketitle}{These authors contributed equally.}{Equal contribution.}{}{}
\makeatother

\title{Grounded Joint-Attention Other-Play for Zero-Shot Coordination}
\author{
    Giulia Benintendi\textsuperscript{\rm 1}\equalcontrib\thanks{Work done while interning at the University of Stuttgart.},
    Constantin Ruhdorfer\textsuperscript{\rm 2}\equalcontrib,
    Fabian K\"ogel\textsuperscript{\rm 2},
    Andreas Bulling\textsuperscript{\rm 2}
}
\affiliations{
    \textsuperscript{\rm 1}University of Zurich\\
    \textsuperscript{\rm 2}University of Stuttgart\\
    giulia.benintendi@uzh.ch
}

\begin{document}

\maketitle

\begin{abstract}
Joint attention -- the human ability to share a common visual or cognitive focus with others -- enables a \emph{meeting of minds} that lets us coordinate even with unfamiliar partners. 
In this work we investigate whether equipping AI agents with a similar mechanism can enable such zero-shot coordination.
We introduce \textbf{M}utual \textbf{A}ttention for zero-shot \textbf{TE}aming (MATE): a novel multi-agent reinforcement learning method inspired by human joint attention.
MATE encourages agents to coordinate their actions by aligning their visual attention on scene-salient objects during the interaction rather than relying on arbitrary partner-dependent conventions established during training.
Unlike symmetry-breaking approaches that merely prevent brittle conventions from emerging, MATE actively promotes coordination through an environment-grounded signal that is naturally shared across partners.
We evaluate MATE on three benchmarks: our Card Alignment Game, designed to isolate brittle convention formation, and the more challenging Level-Based Foraging and OvercookedV2 benchmarks.
Our experiments consistently show that a joint-attention-inspired signal improves coordination with unknown partners, underlining MATE's potential as 
a general coordination mechanism that complements and surpasses symmetry-breaking approaches.

\end{abstract}

\section{Introduction}
\begin{figure}[htbp]
  \centering
  \includegraphics[width=\linewidth]{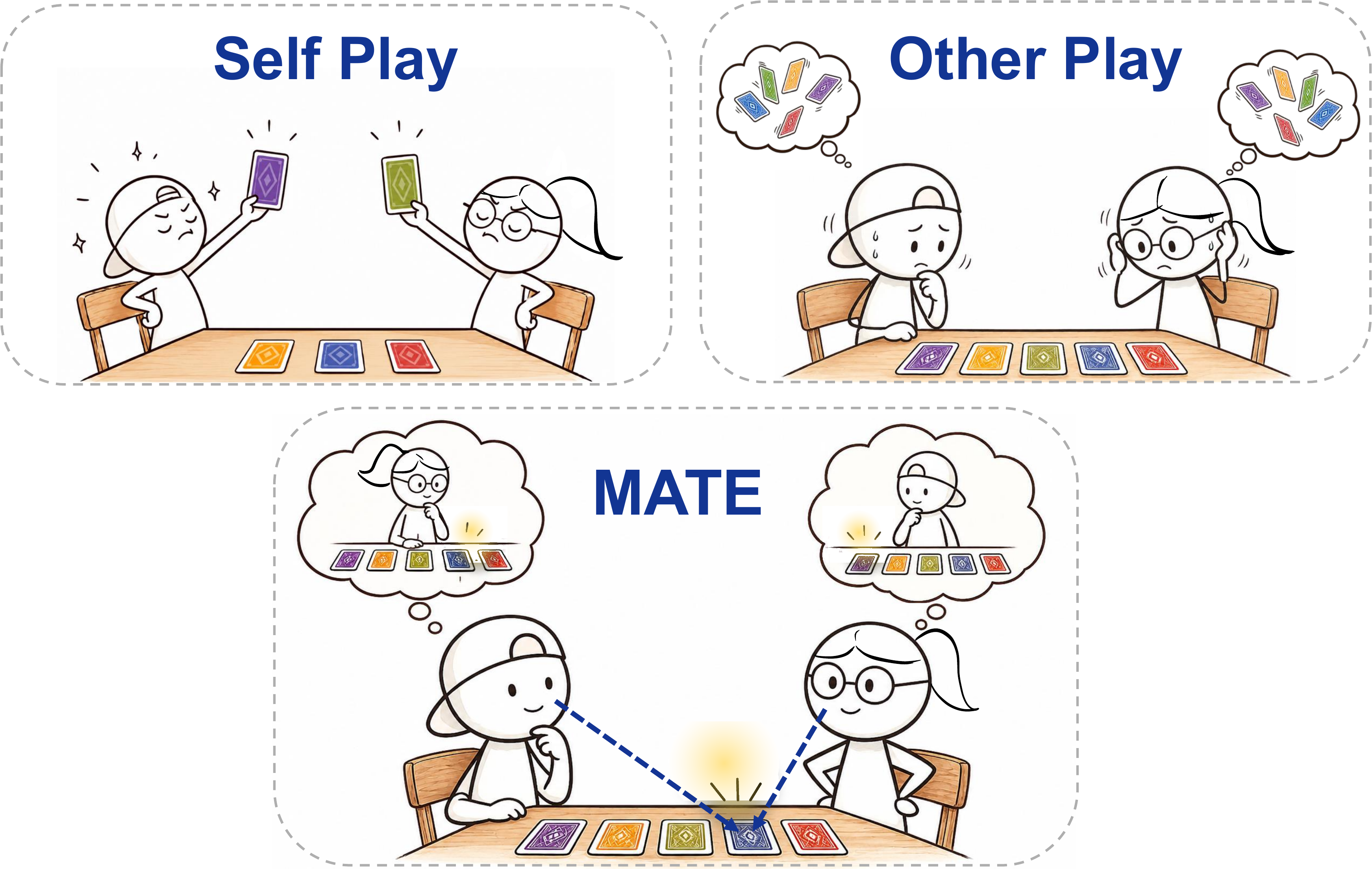}
  \caption{
  Zero-Shot Coordination using joint attention. Alice \bottomagent{} and Bob \topagent{}, trained independently, must pick the same card without explicit communication. \textit{Self-play} agents commit to arbitrary conventions, \textit{other-play} agents have no signal at all to coordinate on, while \textit{MATE} agents read each other's attention and converge on a shared choice.
  }
  \label{fig:teaser}
\end{figure}
One of the most common approaches for training agents in multi-agent reinforcement learning (MARL) is \emph{self-play} (SP)~\cite{tesauro95selfplay}, in which agents learn by interacting with a copy of themselves. 
Despite yielding strong results in competitive~\cite{silver2016alphago,berner2019dota,vinyals2019alphastar,baker2019emergent} and cooperative settings~\cite{dewitt2020multiagentcommonknowledgereinforcement,hu2021simplifiedactiondecoderdeep,yu2022surprisingeffectivenessppocooperative}, agents trained via SP struggle to generalise to novel partners~\cite{carroll2020utilitylearninghumanshumanai,hu2021otherplayzeroshotcoordination,strouse2021collaborating}.
This motivates research on \emph{zero-shot coordination} (ZSC): coordinating with unseen partners without prior coordination~\cite{stone2010aht,hu2021otherplayzeroshotcoordination}.
ZSC performance is usually evaluated in \emph{cross-play} (XP) by pairing agents from independent training runs. 
Performance degrades substantially~\cite{hu2021otherplayzeroshotcoordination} when agents approach the same task through incompatible arbitrary conventions that break environment symmetries during training~\cite{foerster2019bayesianactiondecoderdeep}.
Consider two players tasked with picking the same card out of five without explicit communication (see Figure~\ref{fig:teaser}).
Within a training run, two agents may converge on a shared convention, such as ``always pick green'', while an agent from a different run may have learned to ``always pick purple''.
When paired, neither agent knows which convention the other learned, and they fail to coordinate.
Solving ZSC thus requires coordination signals that a novel partner can independently recover, rather than conventions tied to a %
training run.
To solve this issue, humans use joint attention as a social signal to dynamically resolve ambiguity and establish the relevant object of joint action~\cite{tomasello1995joint,carpenter2013jointattention}.
In the card example, humans would look at their intended card and %
track their partner's gaze to successively converge on a joint choice.
This succeeds even between strangers because human
joint attention generalises across partners, providing a cue that is produced within the ongoing interaction and that can be interpreted in context.
We hypothesise that \emph{joint attention} could play a similar role for AI agents and ask:
\begin{quote}
\emph{Can joint attention serve as a coordination signal for independently trained partners at test time?}
\end{quote}
We propose \textbf{M}utual \textbf{A}ttention for zero-shot \textbf{TE}aming (MATE), a novel multi-agent reinforcement learning method inspired by human visual joint attention.
Each agent receives its partner's attention as a coordination signal and conditions its policy on it, making attention interactive: Agents do not merely learn to align their attention on the same object, but also to use the partner's visual focus to guide their own attention and actions as the interaction unfolds (see Figure~\ref{fig:teaser}).
We build MATE on top of \emph{other-play} (OP)~\cite{hu2021otherplayzeroshotcoordination} that targets brittle convention formation by exploiting known environment symmetries:
Rather than coordinating with an identical copy, an agent is trained with a randomly symmetry-permuted version of itself, preventing convergence on arbitrary conventions.
In the card example in Figure~\ref{fig:teaser}, OP can prevent run-specific choices, such as always selecting purple or green, but it does not determine which card agents should choose:
While OP removes one source of failure, it does not provide any within-episode signal for selecting one among several equally plausible cards.
MATE supplies this missing signal through the partner's visual attention, allowing agents to coordinate where OP alone cannot.

We evaluate MATE first in our \emph{Card Alignment Game} introduced above, which isolates this coordination task in its simplest form, and then in two established coordination benchmarks: Level-Based Foraging ~\cite{albrecht2015gametheoreticmodelbestresponselearning} and OvercookedV2 ~\cite{gessler2025overcookedv2rethinkingovercookedzeroshot}.
Our results suggest that successful ZSC may require not only preventing incompatible arbitrary conventions, but also enabling agents to negotiate coordination during execution by establishing a shared attentional focus.
This positions MATE as a method for fostering generalisable coordination strategies that transfer to novel partners.
Our contributions are:
\begin{enumerate}    
    \item We introduce MATE, a joint attention training mechanism built on top of OP, in which each agent observes its partner's visual attention and learns to act on it. 
    \item We identify coordination problems in which enforcing symmetry-equivariant policies alone is insufficient, and introduce the Card Alignment Game as a minimal testbed that isolates this limitation.
    \item We show that MATE outperforms competitive baselines in the Card Alignment Game as well as Level-Based Foraging and OvercookedV2, together with qualitative evidence that coordination stems from mutual attention rather than brittle conventions.
\end{enumerate}

\section{Related Work}
\subsection{Zero-Shot Coordination}
ZSC investigates the ability of agents to coordinate with independently trained partners at test time, without co-adaptation.
This is challenging when a task admits multiple equivalent but mutually incompatible optimal joint policies.
We focus on this setting rather than broader ad-hoc teamwork (AHT), where agents collaborate with previously unknown partners whose behaviours may initially be unknown~\cite{stone2010aht}, as AHT is commonly treated as a natural next step once ZSC performance improves~\cite{dizdarevic2025adhoc,ruhdorfer2026yokailearningenvironmenttracking}.
Existing ZSC methods primarily address this coordination problem by discouraging arbitrary symmetry breaking during training.
OP~\cite{hu2021otherplayzeroshotcoordination} independently transforms each agent’s Dec-POMDP using sampled environment relabellings, preventing the arbitrary conventions that SP would otherwise form around environment symmetries.
However, its performance is bounded by the symmetries the environment admits.
Off-Belief Learning~\cite{hu2021offbelieflearning} on the other hand is an algorithm that prevents arbitrary conventions and improves ZSC by taking as input a reference policy $\pi_0$ and optimizes a new unique policy $\pi_1$ under the assumption that previous actions were taken by $\pi_0$ while future actions will be taken by $\pi_1$ using a belief model.
More recently, \citet{forkel2026highentropy} showed that high entropy regularisation promotes symmetry-equivariant and XP-compatible policies, with High-Entropy (HE) IPPO, a standard IPPO with an unusually large entropy coefficient, achieving state-of-the-art inter-seed XP
score in Hanabi.
These approaches share the premise that coordination can be recovered by suppressing arbitrary conventions. 
This is not always sufficient.
When several possible referents remain symmetric within the current episode, removing arbitrary conventions does not by itself determine what the agents should coordinate around.
We instead equip agents with a coordination signal grounded in the episode itself, namely shared attention, that remains informative even when no symmetry is left to exploit.

\subsection{Joint Attention as a Coordination Signal}
Joint attention is the shared and mutually recognised focus of two individuals on the same object or event.
It is a foundational mechanism in human social cognition~\cite{tomasello1995joint,mundy2007joint,carpenter2013jointattention} and has been studied for coordination in human-robot interaction~\cite{Staudte2011HRI, anzalone2019humanactivities,martinez2024HRI}.
In MARL, \citet{lee2021jointattentionmultiagentcoordination} equip agents with spatial attention heads and an intrinsic reward that penalises divergence between their attention maps, encouraging them to attend to the same elements of a shared observation; this improves SP coordination in gridworld tasks.
However, because their evaluation is limited to SP, the shared-attention incentive may act as an additional convention-forming channel: co-trained agents can simply learn to align their attention on an arbitrary, possibly task-irrelevant location to initiate a coordinated behaviour, without producing a signal that transfers to independently trained partners.
To leverage communication, prior work in MARL learns protocols for exchanging task-relevant information~\cite{Foerster2016learning,Sukhbaatar2016learning,Mordatch2018emergence}, but these often fail to transfer to independently trained partners~\cite{Bullard2020exploring,Tucker2021emergent}.
MATE instead uses spatial attention as an environment-grounded communication channel.
Without a pre-established convention, two factors matter beyond aligned attention: agents need to be mutually aware that they share a focus~\cite{carpenter2013jointattention} and the focus needs to be on a relevant object~\cite{csibra2009pedagogy}.
MATE addresses both limitations by enforcing attention to be not only aligned but also focused on objects, yielding an observable signal for establishing an episode-specific target with novel partners.

\section{Preliminaries}
\label{sec:preliminaries}

\subsection{Decentralised POMDPs}
We model the cooperative setting as a decentralised partially observable Markov
decision process (Dec-POMDP)~\cite{oliehoek2007decpomdp}, a tuple
$\left(\mathcal{S}, n, \{\mathcal{A}^i\}_{i=1}^n,\{\mathcal{O}^i\}_{i=1}^n, \mathcal{T}, \mathcal{R},\{U^i\}_{i=1}^n, T, \gamma, b_0\right)$, with finite state space $\mathcal{S}$, $n$ agents, local action and observation
spaces $\mathcal{A}^i$ and $\mathcal{O}^i$, transition kernel
$\mathcal{T}(s_{t+1}\mid s_t, a_t)$, shared reward
$\mathcal{R}(s_{t+1}, a_t)$, observation functions $o^i_t = U^i(s_t)$,
horizon $T$, reward discount factor $\gamma$, and initial state distribution $b_0$.
Each
agent $i$ acts on its local action-observation history $\tau^i_t = (o^i_0, a^i_0, \dots, a^i_{t-1}, o^i_t)$ through a local policy $\pi^i(a^i_t \mid \tau^i_t)$.
The joint policy then
factorises as
\begin{equation}
\pi(a_t \mid \tau_t)
=
\prod_{i=1}^n \pi^i(a^i_t \mid \tau^i_t).
\end{equation}
The SP objective is the expected return
\begin{equation}
J_{\mathrm{SP}}(\pi)
=
\mathbb{E}_{\tau_T \sim \pi}
\left[
\sum_{t=0}^{T-1}
\gamma^t\, \mathcal{R}(s_{t+1}, a_t)
\right].
\end{equation}

\subsection{Zero-Shot Coordination}
\label{sec:zsc}
A Dec-POMDP may admit multiple joint policies with high SP return that are nevertheless mutually incompatible~\cite{hu2021otherplayzeroshotcoordination}.
We thus measure the ZSC score through XP, pairing agents trained in different runs.
Given a Dec-POMDP with $n$ agents and $n$ independently trained joint policies $\pi_1,\dots,\pi_n$, we follow prior work~\cite{hu2021otherplayzeroshotcoordination,forkel2026highentropy} and define the XP score between them as
\begin{equation}
J_{\mathrm{XP}}(\pi_1,\ldots,\pi_n)
:= \mathbb{E}_{\phi\sim\mathrm{Perm}(n)}
J\bigl(\pi_{\phi(1)}^1,\ldots,\pi_{\phi(n)}^n\bigr).
\end{equation}
where $\mathrm{Perm}(n)$ is the set of permutations of
$\{1,\dots,n\}$, and $\pi_j^i$ denotes the local policy for agent slot $i$ from joint policy $\pi_j$.
As in this work we consider two-player Dec-POMDPs, this reduces to
\(J_{XP}(\pi_1, \pi_2) = \tfrac{1}{2}\left[ J(\pi_1^1, \pi_2^2) + J(\pi_2^1, \pi_1^2) \right]\).
The goal of ZSC is to find training algorithms that,
when run independently, produce compatible joint policies. Following prior work, we focus on the inter-seed XP setting, i.e. independent training runs vary only in their random seed. %

\subsection{Symmetries in Dec-POMDPs}
A Dec-POMDP symmetry~\cite{hu2021otherplayzeroshotcoordination,forkel2026highentropy} is a tuple of bijections
\begin{equation}
\phi
=
\left(
\phi_{\mathcal{S}},
(\phi_{\mathcal{A}^i})_{i=1}^n,
(\phi_{\mathcal{O}^i})_{i=1}^n
\right)
\end{equation}
over $\mathcal{S}$, $\mathcal{A}^i$, and $\mathcal{O}^i$ while leaving $\mathcal{T}$, $\mathcal{R}$, and $\{U^i\}_{i=1}^n$ invariant.
We write $\Phi$ for the set of
all such symmetries. A symmetry acts on a joint policy by relabelling its inputs
and outputs,
\begin{equation}
\phi(\pi^i)(a^i \mid \tau^i)
:=
\pi^i
\left(
\phi^{-1}(a^i)
\mid
\phi^{-1}(\tau^i)
\right),
\end{equation}
where $\phi^{-1}(\tau^i)$ denotes the component-wise relabelling of the
actions and observations in the action-observation history. A joint policy
$\pi$ is symmetry equivariant if $\phi(\pi)=\pi$ for all
$\phi \in \Phi$.
\subsection{Other-play}
OP addresses brittle conventions by optimising SP under independently sampled Dec-POMDP symmetries~\cite{hu2021otherplayzeroshotcoordination}. 
For each agent $i$, a symmetry $\phi^i \sim \Phi$ is drawn per episode, and OP maximises
\begin{equation}
J_{\text{OP}}(\pi) = \mathbb{E}_{(\phi^1,\dots,\phi^n)\sim\Phi^n}
\bigl[J_{\text{SP}}\bigl(\phi^1(\pi^1),\dots,\phi^n(\pi^n)\bigr) \bigr]
\end{equation}
This promotes robustness to relabellings and discourages arbitrary symmetry breaking, yet
$\Phi$ must be known a priori.

\subsection{Joint Attention for Social Learning}
\citet{lee2021jointattentionmultiagentcoordination} introduced a computational analogue of joint attention that improved $J_{SP}$ and sample efficiency across several gridworld tasks.
For each agent $k$, a single $h \times w$ spatial attention map $A_t^k$ is obtained by averaging the visual attention maps of the $m$ heads.
Joint attention is modelled as the Jensen-Shannon divergence (JSD) between each pair of agents' attention maps, using an auxiliary reward term:
\begin{equation}
    r_t^{\mathrm{JA}} = -\sum_{j=1}^{n} \sum_{i=1}^{n} \mathrm{JSD}\!\left(A_t^i \,\|\, A_t^j\right).
\end{equation}
Each agent's objective combines the environment reward with the shared joint-attention intrinsic motivation:
\begin{equation}
    J(\pi^i) = \mathbb{E}_\pi\!\left[\, \sum_{t=0}^{\infty} \gamma^t \left(r_{t+1}^i + \beta\, r_t^{\mathrm{JA}}\right) \;\middle|\; s_0 \right],
\end{equation}
where $\beta$ controls the weight of the joint-attention objective.

\section{Mutual Attention for Zero-Shot Teaming}
\label{sec:method}

We extend OP and the spatial-attention architecture of \citet{lee2021jointattentionmultiagentcoordination} by additionally grounding the attention in the environment.
We call this method \emph{MATE}, which equips OP-trained IPPO agents~\cite{dewitt2020ippo,schulman2017ppo} with a model of joint attention.
Each agent shares its focus with its partner and learns to steer its own attention toward the partner's future focus and on task-relevant objects.

\subsection{Sharing the partner's attention}
\label{sec:sharing}
At every step, agent $i$ receives its partner's previous attention map $A^j_{t-1}$ as a channel appended to its image observation.
Because OP expresses each agent's observation in a different permuted frame, we re-express the partner's map in $i$'s frame by undoing $j$'s symmetry and applying $i$'s:
\begin{equation}
\tilde{A}_{t-1}^{j} := \left(\phi_A^i \circ (\phi_A^j)^{-1}\right) A_{t-1}^{j},
\end{equation}
where $\smash{\phi^j_A}$ denotes the action of $\smash{\phi^j}$ on spatial attention maps.

\subsection{Grounding attention in task-relevant objects}
Parallel attention is insufficient for joint attention, as both agents must intentionally share a focus on a relevant target to support coordinated action~\cite{carpenter2013jointattention,csibra2009pedagogy}.
Alignment on an irrelevant background patch contributes nothing to the task unless it becomes another arbitrary convention, whereas attention to task-relevant objects provides a meaningful coordination signal that novel partners can interpret within the episode.
We therefore add an auxiliary loss that grounds attention and encourages convergence on task-relevant entities, aligning attention at the \emph{object} level.
Because each agent observes the partner's previous attention map, this does not passively align attention maps; rather, it trains agents to intentionally condition their own focus on the partner's. 

We assume a training-time object detector $\mathrm{OBJ}$ that returns the task-relevant objects $\mathcal{E}_t = \mathrm{OBJ}(o_t)$ at step $t$, each object $e$ covering a region $\mathcal{R}(e)$ of the attention grid. 
Mapping agent $i$'s attention back to the shared frame, $\bar{A}_t^i := (\phi_A^i)^{-1} A_t^i$, and summing it over each object's cells, leads to a distribution $p_t^i$ over objects, comparable across agents.
We train $p_t^i$ towards the partner's discounted future focus $q_t^j$, which weights each object by a discount factor $\eta^{\delta_t(e)}$, where $\delta_t(e)$ is the number of steps until partner $j$ next attends object $e$. 
Agent $i$'s attention is matched to this target by cross-entropy, summed over both agents and the horizon,
\begin{equation}
\mathcal{L}_{\mathrm{MATE}}
= -\frac{1}{2T}
\sum_{i=1}^{2}\sum_{t=0}^{T-1}
\sum_{e \in \mathcal{E}_t}
q_t^j(e)\log p_t^i(e).
\end{equation}
The full objective adds this to the IPPO loss,
\begin{equation}
    \mathcal{L} = \mathcal{L}_{\mathrm{IPPO}} + \lambda_{\mathrm{MATE}}\,\mathcal{L}_{\mathrm{MATE}}.
\end{equation}
Algorithm~\ref{alg:mate} presents pseudocode for the two-player case.

\begin{algorithm}[tb]
\caption{MATE}
\label{alg:mate}
\textbf{Require:} policy $\pi_\theta$, symmetry group $\Phi$, detector $\mathrm{OBJ}$, discount $\eta \in [0,1)$, weight $\lambda_{\mathrm{MATE}}$
\begin{algorithmic}[1]
\STATE Sample OP symmetries $\phi^1, \phi^2 \sim \Phi$
\STATE Initialise partner maps $A_{-1}^1, A_{-1}^2$ to uniform
\FOR{$t = 0,\dots,T-1$}
    \FOR{agent $i \in \{1,2\}$, partner $j \neq i$}
        \STATE Reframe partner map $\tilde{A}_{t-1}^{j} = (\phi_A^i \circ (\phi_A^j)^{-1})\, A_{t-1}^j$
        \STATE Sample action and attention $a_t^i, A_t^i \sim \pi_\theta(\phi^i(o_t^i), \tilde{A}_{t-1}^{j})$
    \ENDFOR
    \STATE Step environment with $a_t=(a_t^1,a_t^2)$.
\ENDFOR
\STATE Unmap to unpermuted frame $\bar{A}_t^i = (\phi_A^i)^{-1} A_t^i$
\STATE Sum onto objects $p_t^i(e) \propto \sum_{u \in \mathcal{R}(e)} \bar{A}_t^i(u)$
\FOR{agent $i$, partner $j \neq i$}
    \STATE Find partner's focus $m_t^j = \arg\max_e p_t^j(e)$
    \STATE Build target $q_t^j \gets \eta\, q_{t+1}^j$, $q_t^j(m_t^j) \gets 1$, normalise
\ENDFOR
\STATE Update $\theta$ by PPO on $\mathcal{L} = \mathcal{L}_{\mathrm{IPPO}} + \lambda_{\mathrm{MATE}}\,\mathcal{L}_{\mathrm{MATE}}$.
\STATE \textbf{return} $\theta$
\end{algorithmic}
\end{algorithm}

\section{Experiments}
\label{sec:experiments}
We first test MATE in the Card Alignment Game, a toy environment that we propose to validate the core intuition behind our method in isolation.
We then evaluate MATE in two more complex domains, Level-Based Foraging (LBF) \cite{albrecht2015gametheoreticmodelbestresponselearning} and OvercookedV2 \cite{gessler2025overcookedv2rethinkingovercookedzeroshot}, which require agents to agree on a shared target and infer coordination from their partner's actions.

\subsection{Card Alignment Game}
\label{subsec:card-alignment-game}
\begin{figure}[t]
  \centering
  \includegraphics[width=\linewidth]{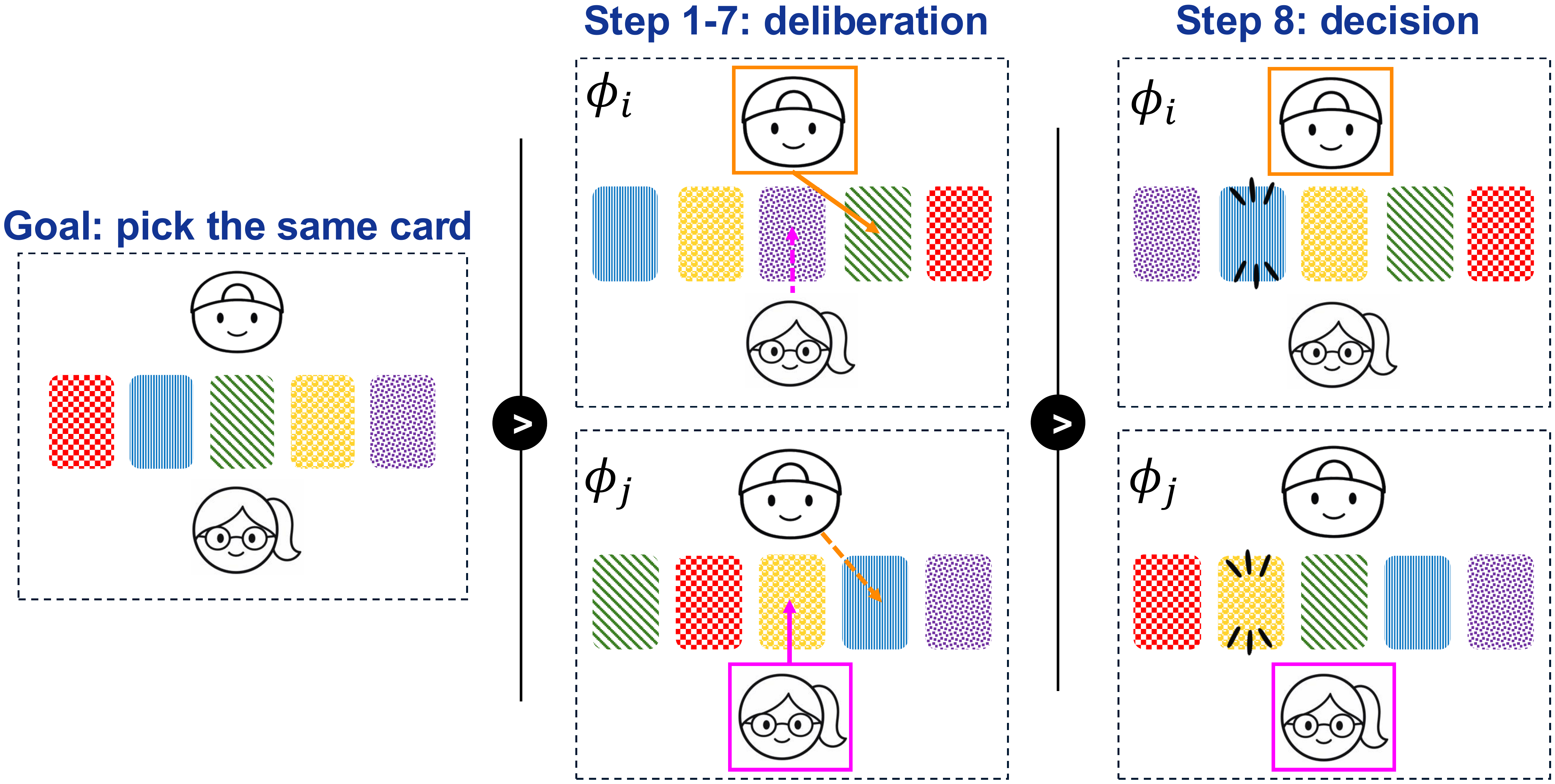}
  \caption{
  Sample episode in the Card Alignment Game. Alice \bottomagent{} and Bob \topagent{} must select the same card in the unpermuted layout (left). To prevent position or colour conventions, they observe differently permuted frames. %
  Finally, both agents choose independently.
    }
  \label{fig:card_game_schema}
\end{figure}
The Card Alignment Game (see Figure~\ref{fig:card_game_schema}) is a cooperative two-player game in which agents must independently pick the same card out of five. Each episode contains seven deliberation steps, during which agents point to cards in their own local frame, followed by one private decision step.
At that step, the pair receives $R = 1$ if both agents pick the same card and $R = 0$ otherwise.
The game is kept simple to isolate whether joint attention alone can provide a sufficiently strong signal to enable coordination. 

\paragraph{Symmetries}
The game admits symmetries over card colours and positions, which SP agents can exploit to form arbitrary conventions, such as always selecting ``red'' or ``left''. 
To avoid this, we employ OP by transforming each agent's observation into $o^i = \phi^i(o)$ where $\phi^i=\phi^i_{\mathrm{col}}\circ\phi^i_{\mathrm{pos}}$ is the composition of random individual position and colour permutations.
Coordination must then be recovered within the episode, rather than through colour- or position-based conventions.
In the example of Figure~\ref{fig:card_game_schema}, Bob \topagent{} sees the unpermuted layout transformed as follows: (i) the position permutation is $\phi^i_{\mathrm{pos}}=(2,3,1,4,5)$, %
and (ii) colour relabelling 
\[
\begin{gathered}
\phi^i_{\mathrm{col}}:
\redcard\!\mapsto\!\purplecard,\,
\bluecard\!\mapsto\!\bluecard,\,
\greencard\!\mapsto\!\yellowcard,
\yellowcard\!\mapsto\!\greencard,\,
\purplecard\!\mapsto\!\redcard.
\end{gathered}
\]
Therefore, Bob \topagent's local observation is
\[
o^i
=
\phi^i(o)
=
(\phi^i_{\mathrm{col}}\circ\phi^i_{\mathrm{pos}})(o)
=
(\bluecard,\yellowcard,\purplecard,\greencard,\redcard).
\]

\paragraph{Observation and action space}
Each agent observes its permuted frame and a scalar indicating whether the episode is in deliberation or at the final decision step.
The action space contains five actions, each corresponding to one of the cards: during deliberation, an action points to a card in the agent's local frame; at the final step, it privately selects that card.
The cards form the object set for the attention loss, with one $\mathcal{R}(e)$ per card.

\begin{figure*}[t]
\centering
\begin{subfigure}[b]{3.52in}
  \centering
  \includegraphics[width=\linewidth]{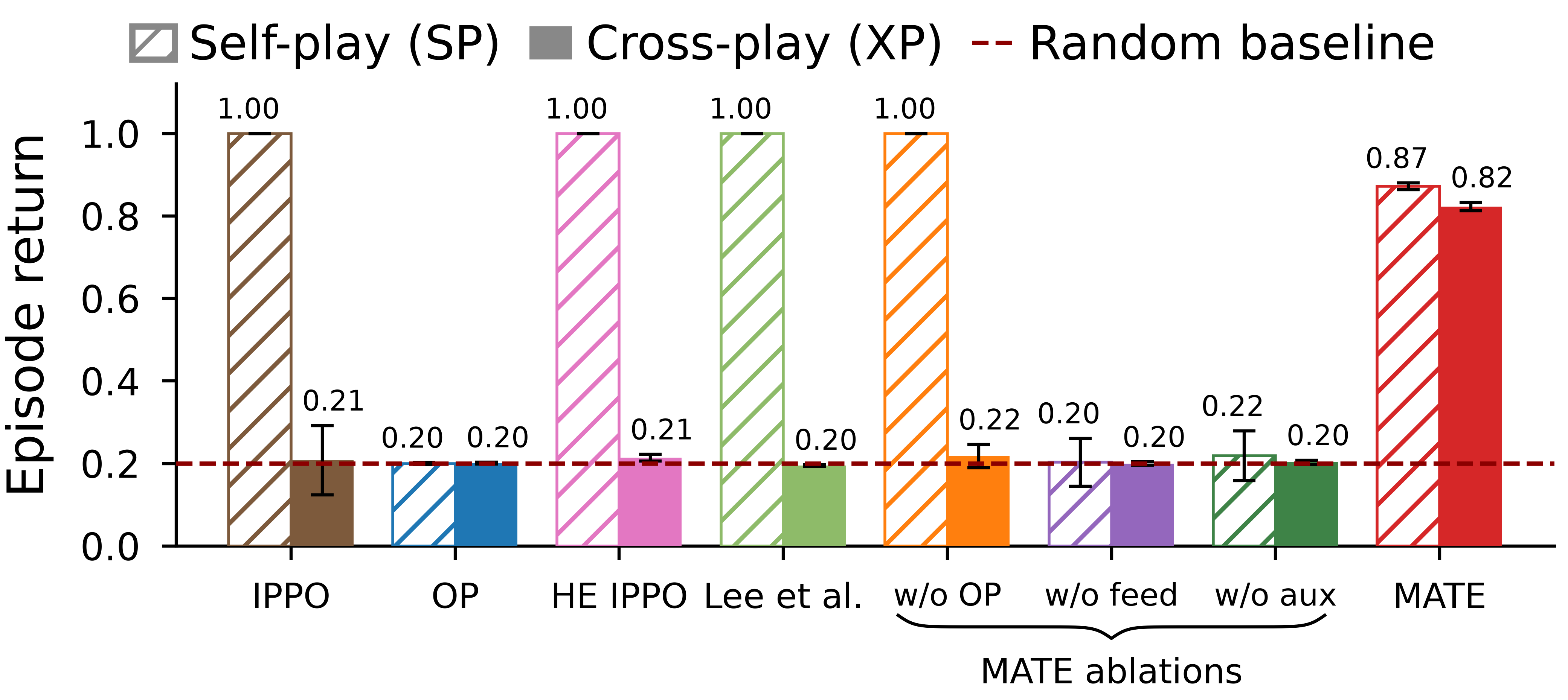}
  \caption{Card Alignment Game}
  \label{fig:results-card-game-bar}
\end{subfigure}\hfill
\begin{subfigure}[b]{1.58in}
  \centering
  \includegraphics[width=\linewidth]{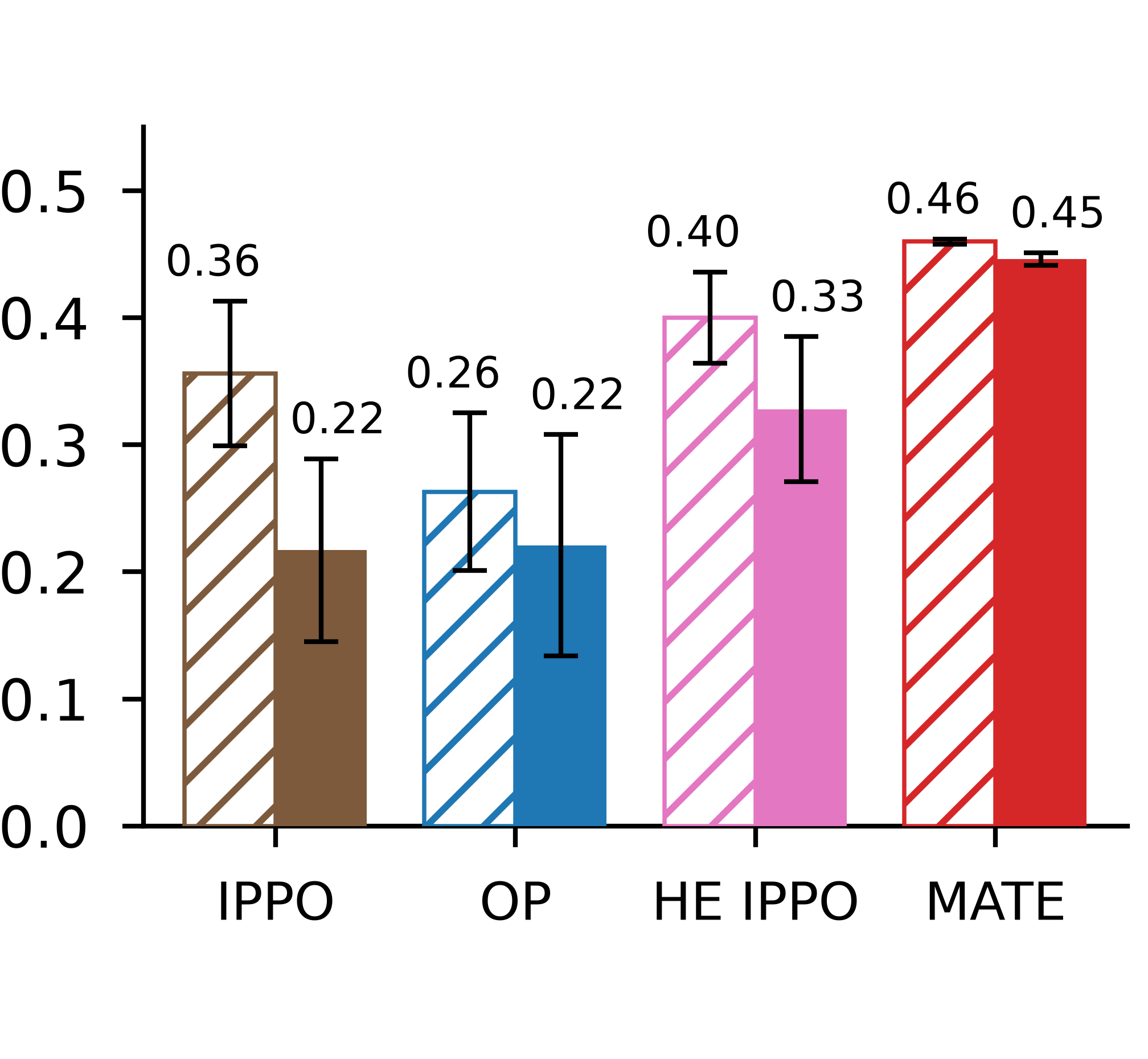}
  \caption{LBF}
  \label{fig:results-LBF-bar}
\end{subfigure}\hfill
\begin{subfigure}[b]{1.89in}
  \centering
  \includegraphics[width=\linewidth]{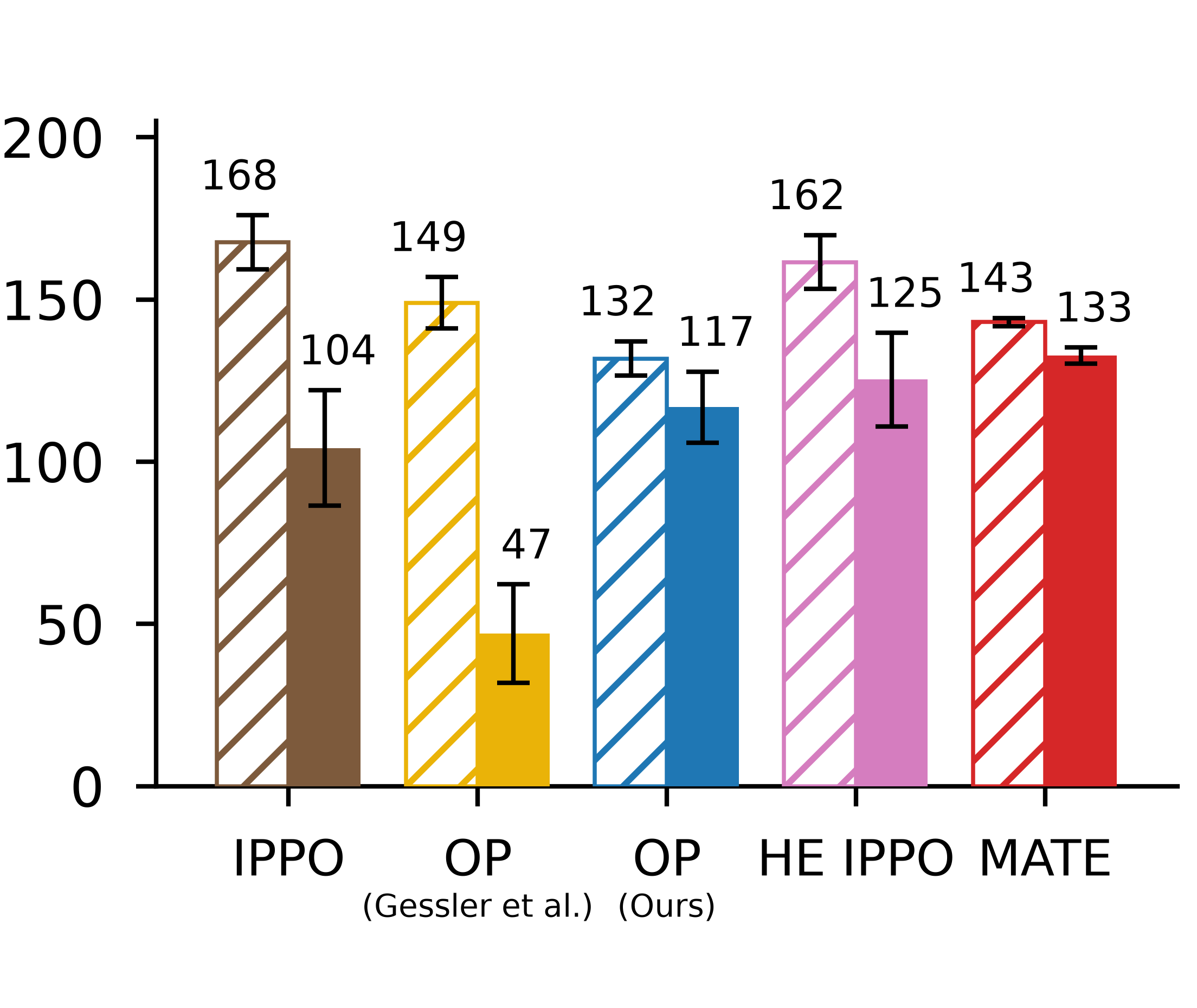}
  \caption{OvercookedV2}
  \label{fig:results-ocv2-bar}
\end{subfigure}
\caption{%
SP (hatched) and XP (solid) scores for the Card Alignment Game (a), LBF (b) and OvercookedV2 (c). Error bars show SEM. MATE achieves the highest XP across all three environments.
}
\label{fig:results-overall}
\end{figure*}

\subsection{Level-Based Foraging}
LBF~\cite{albrecht2015gametheoreticmodelbestresponselearning} is a cooperative grid-world task in which agents navigate to apples and collect them by executing a load action.
We use the forced-coordination setting, where each apple can be collected only if both agents load it jointly.
Experiments are run on a fully observable $12 \times 12$ grid with two agents and eight apples.
We shorten the episode horizon from $100$ to $50$ steps to increase coordination pressure.
Agents receive image observations to test our method in conditions that more naturally resemble real-life deployment scenarios.
Under OP, each observation is independently transformed by a random symmetry of the grid, using horizontal and vertical reflections.
For the object-level attention target, apples are treated as task-relevant objects, with one $\mathcal{R}(e)$ for each apple.

\subsection{OvercookedV2}
OvercookedV2~\cite{gessler2025overcookedv2rethinkingovercookedzeroshot} introduces partial observability and asymmetric information into Overcooked, one of the most widely used ZSC benchmarks.
We evaluate on \emph{Demo Cook Simple}, where only one agent observes the recipe and the other must infer the target dish from its partner's behaviour, particularly the ingredients it carries and places in the pot.
The exploration of additional layouts is left to future work.
This setting makes it a natural test of whether joint attention can support implicit communication through %
partner behaviour.
We extend the JaxMARL~\cite{rutherford2024jaxMARL} environment with image observations and, following \citet{gessler2025overcookedv2rethinkingovercookedzeroshot}, apply OP by independently permuting the two recipe ingredients in each agent's observation.
While \citet{gessler2025overcookedv2rethinkingovercookedzeroshot} embed each local view in a zero-padded global grid, we centre the observation on one agent and crop it to the view radius to reduce dependence on fixed layout coordinates and improve generalisation~\cite{ye2020rotationtranslation}.
Partner attention is shared only within the receiver's visible region, preventing leakage of private recipe information yet making OvercookedV2 particularly challenging for MATE, as attention can provide a coordination signal only intermittently.
Task-relevant kitchen objects, including carried ingredients, form the object set used by the attention loss.

\section{Results}
\label{sec:results}
We evaluate whether MATE improves ZSC by comparing SP and XP return~\cite{hu2021otherplayzeroshotcoordination}, with XP computed from disjoint pairs of independently trained policies following~\citet{forkel2026highentropy}.
Across all environments we compare IPPO, OP, HE IPPO~\cite{forkel2026highentropy} and MATE, additionally including the joint-attention baseline of~\citet{lee2021jointattentionmultiagentcoordination} in the Card Alignment Game. 
Results are averaged over 48 seeds in the Card Alignment Game and 12 in LBF and OvercookedV2.
OP and MATE are trained under the same permutation for a fair comparison; no permutations are applied during evaluation.
All methods are evaluated at the checkpoint with the highest training return.
For HE IPPO, we always report the SP and XP scores at the entropy coefficient achieving the highest XP return in the sweep shown in Figure~\ref{fig:high-entropy}.

\subsection{Card Alignment Game}
In the Card Alignment Game (Figure~\ref{fig:results-card-game-bar}), IPPO solves SP but falls to chance in XP, showing that independently trained agents learn incompatible conventions.
OP rules out brittle conventions, but without a coordination signal, it remains at chance in both SP and XP. HE IPPO also fails: none of the tested entropy coefficients yields XP above chance, and large coefficients collapse non-greedy SP performance (Figure~\ref{fig:high-entropy}).
The \citet{lee2021jointattentionmultiagentcoordination} joint-attention objective behaves similarly: varying $\beta$ over three orders of magnitude leaves the outcome unchanged, with perfect SP but chance XP. 
MATE instead obtains high return in both settings ($0.872 \pm 0.008$ SP, $0.823 \pm 0.010$ XP), moving substantially above chance while keeping the SP-XP gap small.
\begin{figure*}[t]
  \centering
  \includegraphics[
    width=\textwidth
  ]{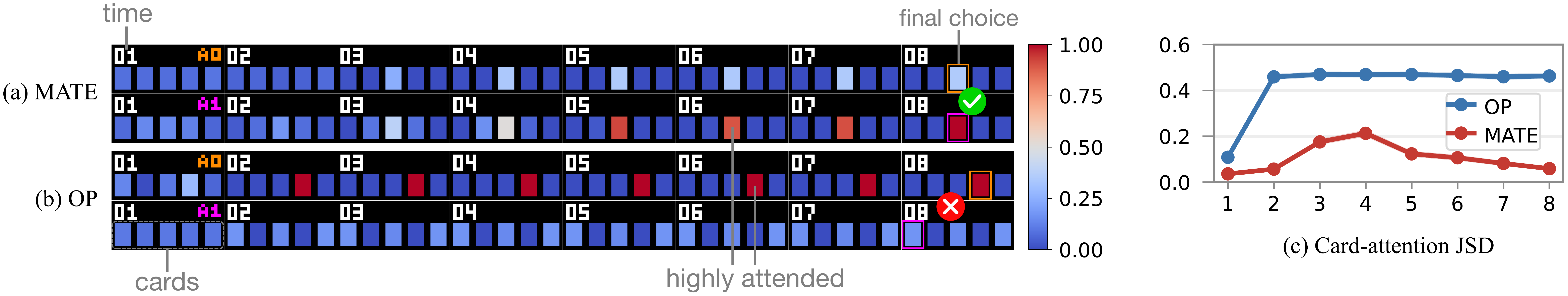}
  \caption{%
    Card Alignment Game attention maps.
    (a) and (b) show per-card attention in the unpermuted frame.
    MATE progressively aligns attention between agents and
    selects the same card, whereas OP maintains divergent attention and chooses different cards. 
    (c) shows card-attention JSD over time; lower values
    indicate stronger alignment.%
  }
  \label{fig:attention-card-game}
\end{figure*}
Figure~\ref{fig:attention-card-game} shows the corresponding attention
patterns. MATE agents progressively align attention on the same card and select it, yielding a strategy that transfers to independently trained partners. Under OP alone, attention remains agent-specific and final choices do not align.

\paragraph{MATE Ablation Study}
We analyse how each component of MATE contributes by individually removing OP (\mbox{w/o OP}), the partner-attention input (\mbox{w/o feed}), and the object-level auxiliary loss (\mbox{w/o aux}); see Figure~\ref{fig:results-card-game-bar}.
\mbox{MATE w/o OP} reintroduces brittle conventions, resulting in perfect SP but chance-level XP, while \mbox{w/o aux} at chance level demonstrates that joint attention is insufficient without grounding:
agents still exchange attention, but an ungrounded signal carries no meaning a novel partner can recover.
\mbox{MATE w/o feed} performs equivalently to OP. 
All three components, symmetry-robust training, interaction-time attention sharing, and object-level grounding, are thus necessary.

\subsection{Level-Based Foraging}
In LBF (Figure~\ref{fig:results-LBF-bar}), IPPO exhibits a clear SP--XP gap, dropping from $0.356 \pm 0.057$ in SP to $0.229 \pm 0.068$ in XP. 
OP reduces this gap but also lowers SP performance, while HE IPPO improves both SP and XP without fully closing the gap. 
MATE performs best in both settings, reaching $0.460 \pm 0.002$ in SP and $0.446 \pm 0.003$ in XP. 

\begin{figure*}[t]
\centering
\begin{subfigure}[b]{0.3888\textwidth}   %
  \includegraphics[width=\linewidth]{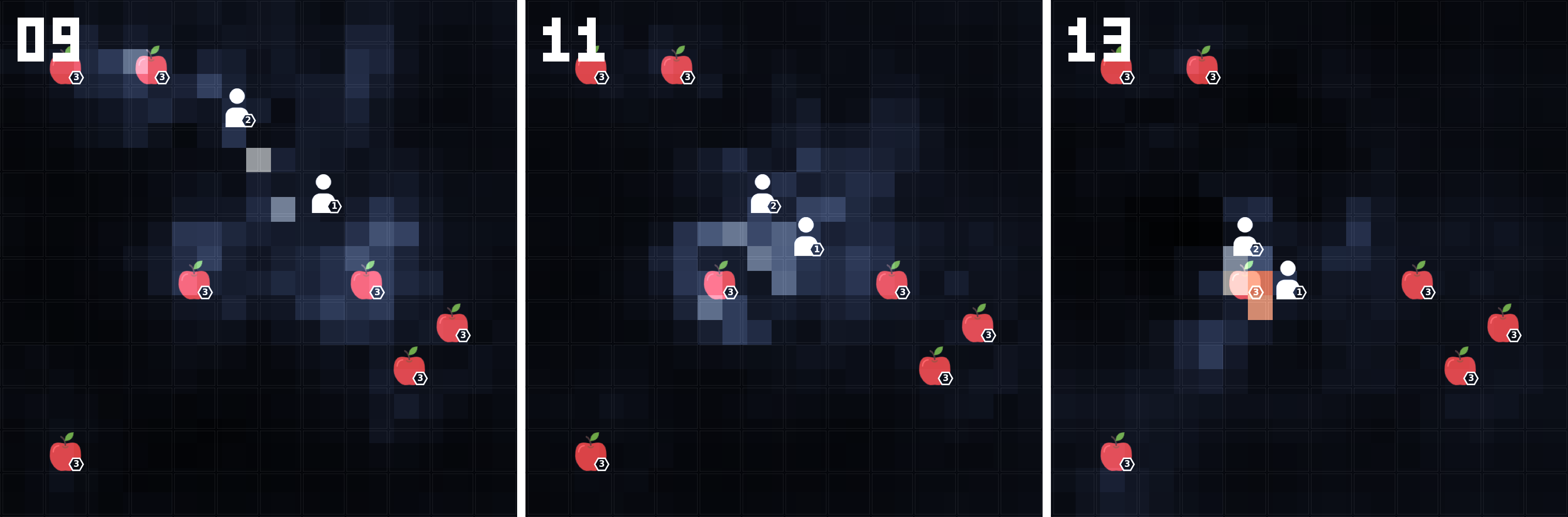}
  \caption{MATE}\label{fig:lbf-mate}
\end{subfigure}\hspace{0.018\textwidth}%
\begin{subfigure}[b]{0.45\textwidth}       %
  \includegraphics[width=\linewidth]{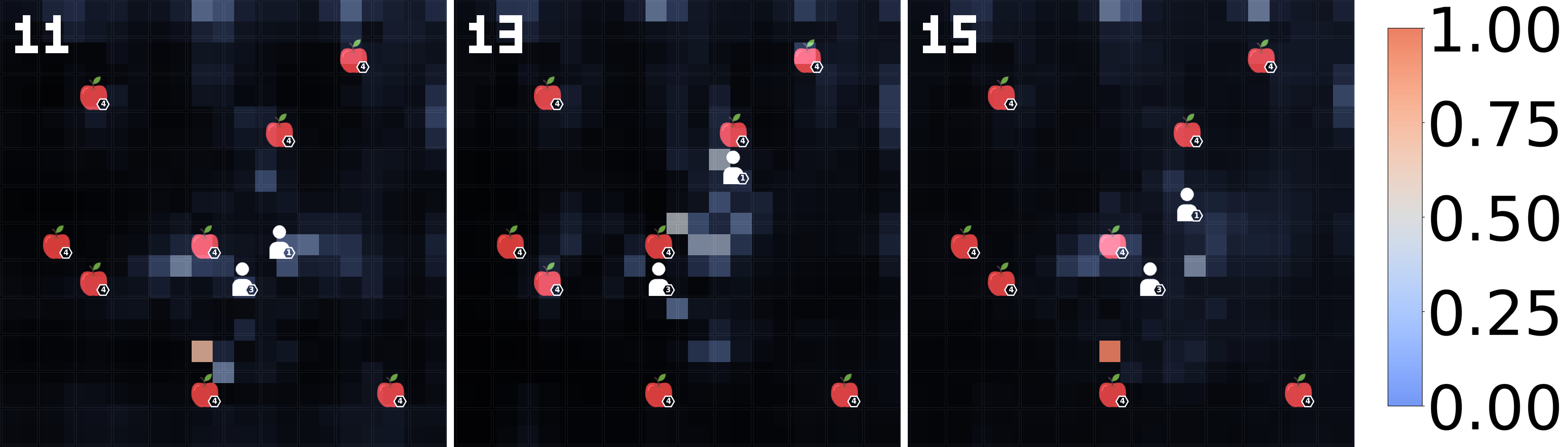}
  \caption{OP}\label{fig:lbf-op}
\end{subfigure}
\caption{
  Level-Based Foraging attention maps.
  (a) and (b) show the combined attention of two independently trained agents in the unpermuted frame. The colour scale indicates normalised attention intensity. When several apples are equally valid targets, MATE agents converge on the same apple, whereas OP agents attend to different regions and fail to establish a common target.}
\label{fig:lbf-attention-maps}
\end{figure*}

Attention maps in Figure~\ref{fig:lbf-attention-maps} show the coordination mechanism. In LBF, several apples are simultaneously equally valid targets and the time budget is short: agents must converge within a few steps, or the episode is lost to wasted moves. MATE partners align their combined attention on the next apple and collect it; OP agents, instead,
each attend to their own vicinity and never commit to a shared target in time.

\subsection{OvercookedV2}
In \textit{Demo Cook Simple} (Figure~\ref{fig:results-ocv2-bar}), MATE achieves the highest XP return, reaching $132.2 \pm 3.0$ in XP and $143.3 \pm 1.4$ in SP.
IPPO drops from $167.6 \pm 8.3$ in SP to $104.2 \pm 17.8$ in XP.
HE IPPO reaches the highest SP return in the sweep ($208.6$), but only $120.6 \pm 51.2$ in XP (see Figure ~\ref{fig:high-entropy}), leaving a substantial SP--XP gap; even its best-XP configuration remains below MATE at $125.3 \pm 14.4$.
OP improves XP to $116.8 \pm 10.9$, considerably above the $47 \pm 15.2$ originally reported for OP by \citet{gessler2025overcookedv2rethinkingovercookedzeroshot}.
In our experiments, this gain comes from replacing zero-padded observations with agent-centred crops: with the original representation, agents develop additional conventions tied to fixed spatial positions, suggesting that ingredient relabelling alone may not capture all relevant symmetries.
We therefore report the original result only as a reference rather than as a directly comparable baseline.

\begin{figure}[t]
  \centering
  \includegraphics[
    width=1\linewidth
  ]{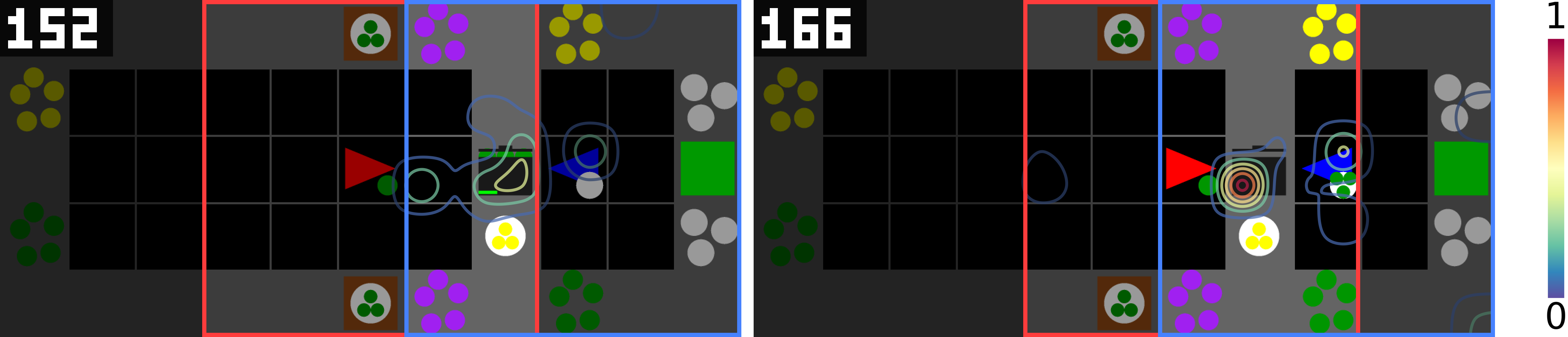}
  \caption{%
  OvercookedV2 attention maps from two moments of interaction between MATE agents.
  Red and blue boxes delimit the agents' respective fields of view; contour lines show their combined attention in the unpermuted frame.
  }
  \label{fig:overcooked-attention}
\end{figure}

The benefit MATE can provide in this layout is inherently constrained.
The agents' fields of view overlap for only $262 \pm 9$ of the $400$ evaluation timesteps and, even then, the shared region covers on average $11 \pm 0.3$ cells and at most $15$ of the $50$ cells in each view, limiting how often mutual attention is effective.
Figure~\ref{fig:overcooked-attention} illustrates one such brief window.
Once the informed agent returns to the pot carrying the required ingredient, the ingredient reveals the hidden recipe, while mutual attention directs the uninformed agent's focus towards this cue, indicating that the prepared soup should be delivered.
Nevertheless, MATE outperforms OP under the same agent-centred observations and ingredient permutations, showing that it can improve coordination even under constrained shared visibility, and extending its benefits to partial observability and asymmetric information.

\section{Discussion and Limitations}

Our results show that MATE supports within-episode coordination across tasks requiring target agreement, as in Card Alignment and LBF, and interpretation of behaviour under partial observability, as in OvercookedV2.
This suggests that some coordination problems cannot be solved by suppressing arbitrary conventions alone; for these, grounding coordination in signals recoverable within the episode offers an effective route to generalisation in cooperative multi-agent systems.
A limitation is that MATE assumes access to a detector for task-relevant objects during training.
We also evaluate only with AI partners; whether the learned attention signal stays interpretable and useful in human--AI coordination remains open.

\section{Conclusion}
We introduced MATE, a multi-agent reinforcement learning method that models joint attention as a general zero-shot coordination mechanism between independently trained partners.
Across three diverse environments, MATE closes the SP-XP gap where prior zero-shot coordination methods remain at chance level.
Future work could extend this idea beyond shared focus, using attention-based objectives and attention-avoidance to support richer forms of coordination, including division of labour, help-seeking, warning, and complementary attention to different task-relevant objects.

\appendix
\section{Training Details}
\begin{figure}[!b]
  \centering
  \begin{subfigure}{\linewidth}
    \centering
    \includegraphics[width=0.95\linewidth]{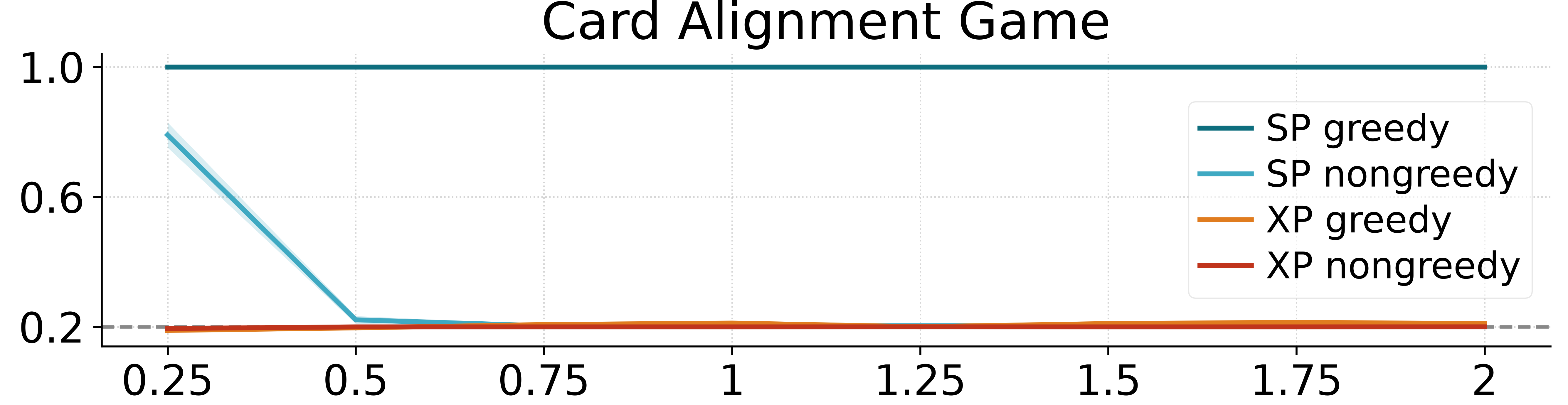}
    \label{fig:high-entropy-card}
  \end{subfigure}
  \begin{subfigure}{\linewidth}
    \centering
    \includegraphics[width=0.95\linewidth]{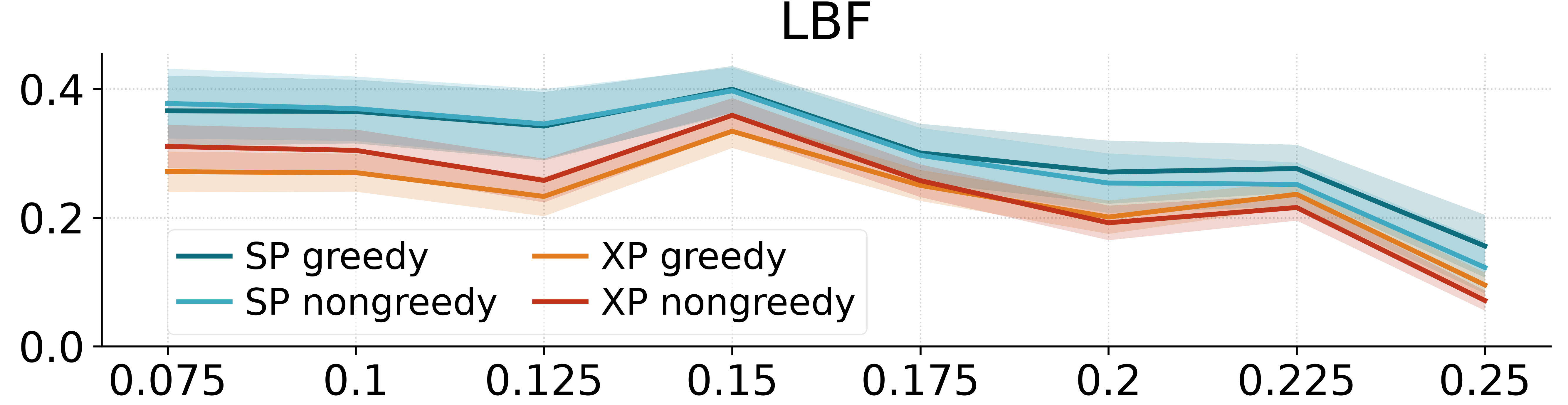}
    \label{fig:high-entropy-LBF}
  \end{subfigure}
  \begin{subfigure}{\linewidth}
    \centering
    \includegraphics[width=0.95\linewidth]{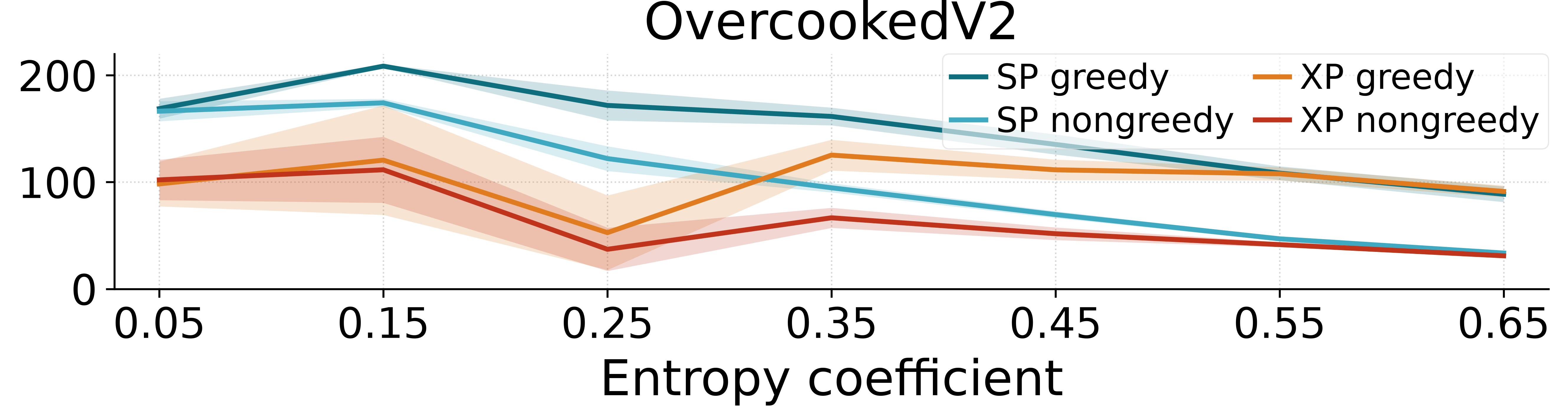}
    \label{fig:high-entropy-ocv2}
  \end{subfigure}
  \caption{
  Scores of HE IPPO for SP and XP, against the entropy coefficient. 
  Shaded regions show SEM.
  }
  \label{fig:high-entropy}
\end{figure}

\subsection{Network Architecture}
All methods use the same network architecture, inspired by the visual-attention architecture of \citet{lee2021jointattentionmultiagentcoordination}.
Since our observations are image-based rather than symbolic grid states, we replace their single convolutional encoder with a residual CNN.
Observations are processed by a four-block residual CNN (\(32\) filters, \(3{\times}3\) kernels, stride \(2\), ReLU~\cite{nair2010relu}), followed by an LSTM of hidden size \(128\) and two fully connected layers of size \(256\) for both actor and critic.

\subsection{Hyperparameters}
Our codebase builds on JaxAHT~\cite{jaxaht2025}, and we ran our experiments on Ubuntu servers equipped with NVIDIA Tesla V100-SXM2 GPUs with 32 GB of memory.
For all methods, we use seed 42 as a base and either incremented it sequentially (OvercookedV2) or split new random seeds using JAX (LBF, Card Alignment Game). 
We tuned PPO hyperparameters per environment from the JaxMARL baselines~\cite{rutherford2024jaxMARL}, \citet{forkel2026highentropy}, and, for OvercookedV2, \citet{gessler2025overcookedv2rethinkingovercookedzeroshot}.
We varied one parameter at a time and retained the best SP-performing configuration.
We swept $\lambda_{\mathrm{MATE}} \in \{10^{-3},\allowbreak\,
3{\cdot}10^{-3},\allowbreak\, 10^{-2},\allowbreak\,
3{\cdot}10^{-2},\allowbreak\, 0.1,\allowbreak\, \mathbf{0.25},\allowbreak\,
0.4,\allowbreak\, 0.5,\allowbreak\, 1,\allowbreak\, 2\}$
for the Card Alignment Game and
$\{\mathbf{10^{-4}},\allowbreak\, 5{\cdot}10^{-4},\allowbreak\,
5{\cdot}10^{-3},\allowbreak\, 5{\cdot}10^{-2}\}$
for LBF.
For the \citet{lee2021jointattentionmultiagentcoordination} baseline, we swept
$\beta \in \{\mathbf{3{\times}10^{-4}},\allowbreak\, 10^{-3},\allowbreak\,
3{\times}10^{-3},\allowbreak\, 10^{-2},\allowbreak\,
3{\times}10^{-2},\allowbreak\, 10^{-1},\allowbreak\, 0.3,\allowbreak\, 1\}$.

\begin{table}[t]
\centering
{%
\small
\setlength{\tabcolsep}{1mm}
\begin{tabular}{@{}lccc@{}}
\toprule
\textbf{Hyperparameter}
    & \textbf{Card Align.}
    & \textbf{LBF}
    & \textbf{OCv2} \\
\midrule
LR
    & $\{3,\mathbf{5},8\}\!\times\!10^{-4}$
    & $4\!\times\!10^{-4}$
    & $\{2.5,\mathbf{4}\}\!\times\!10^{-4}$ \\
LR annealing
    & False
    & True
    & True \\
Steps
    & $5\!\times\!10^{6}$
    & $3\!\times\!10^{6}$
    & $3\!\times\!10^{7}$ \\
Parallel envs
    & $\{512,\mathbf{1024}\}$
    & $\{\mathbf{128},256\}$
    & $\{\mathbf{192},256\}$ \\
Rollout
    & $8$
    & $128$
    & $256$ \\
Epochs
    & $4$
    &
    & \\
Minibatches
    & $16$
    & $32$
    & $\{\mathbf{16},64\}$ \\
Clip $\epsilon$
    & $0.2$
    &
    & \\
Entropy coef.
    & $0.01$
    & $0.01$
    & $\{0.01,\mathbf{0.02}\}$ \\
Value coef.
    & $0.5$
    &
    & \\
GAE $\lambda$
    & $\{.8,.9,\mathbf{.95}\}$
    & $.8$
    & $\{\textbf{.8},.9,.95\}$ \\
Discount $\gamma$
    & $.99$
    &
    & \\
Grad norm
    & $.5$
    &
    & \\
\midrule
\multicolumn{4}{@{}l}{\textit{MATE parameters}} \\
\midrule
$\lambda_{\mathrm{MATE}}$
    & $.25$
    & $10^{-4}$
    & $\{10^{-3},\mathbf{10^{-4}}\}$ \\
$\eta$
    & $.9$
    & $\{.8,\mathbf{.95}\}$
    & $\{.8,\mathbf{.95}\}$ \\
\bottomrule
\end{tabular}%
}
\caption{Training hyperparameters for Card Alignment Game,
LBF, and OvercookedV2. Braces show the values considered during tuning; bold values were selected. Empty entries inherit Card
Alignment Game values.}
\label{tab:hyperparams}
\end{table}

\subsection{High-Entropy IPPO Baseline}
Following~\citet{forkel2026highentropy}, we keep standard IPPO training unchanged and tune the entropy coefficient over a %
larger range (Figure~\ref{fig:high-entropy}).
This should make independently trained policies compatible, with XP matching SP.
In the Card Alignment Game, we extend the sweep to $10^5$, yet the greedy SP--XP gap remains open throughout.
In LBF, greedy SP and XP become close for several coefficients, but never intersect, with SP consistently higher.
In OvercookedV2, greedy XP approaches and eventually exceeds SP only after increasing entropy has substantially degraded SP performance.

\section*{Acknowledgments}
The authors thank the International Max Planck Research School for Intelligent Systems (IMPRS-IS) for
supporting C. Ruhdorfer and F. Kögel.

\section*{Contribution Statement}
G.B. and C.R. jointly developed the method, with C.R. conceiving the initial research direction and method, while G.B. led its implementation and experimental development.
G.B. conducted the experiments, analyzed the experimental results, and led the writing of the paper.
C.R. provided technical leadership, contributed to the experimental analysis, and contributed to the writing of the paper.
F.K. advised on key research decisions and contributed to the writing of the paper.
A.B. supervised the research, acquired funding, and contributed to the writing of the paper.

\bibliography{aaai2027}

\end{document}